\documentclass{llncs}
\usepackage{amsmath}
\usepackage{graphicx}
\usepackage{float}
\usepackage{xcolor}

\begin{document}

\title{A graph-embedded deep feedforward network for disease outcome classification and feature selection using gene expression data}
\titlerunning{graph-embedded deep feedforward networks}  
%
\author{Yunchuan Kong\inst{1} \and Tianwei Yu\inst{1}}
\authorrunning{Kong and Yu} 
%
%
\institute{Department of Biostatistics and Bioinformatics, Emory University, 1518 Clifton Rd, Atlanta, GA 30322, USA\\
\email{tianwei.yu@emory.edu},
}

\maketitle              

\begin{abstract}
Gene expression data represents a unique challenge in predictive model building, because of the small number of samples $(n)$ compared to the huge amount of features $(p)$. This ``$n<<p$'' property has hampered application of deep learning techniques for disease outcome classification. Sparse learning by incorporating external gene network information could be a potential solution to this issue. Still, the problem is very challenging because (1) there are tens of thousands of features and only hundreds of training samples, (2) the scale-free structure of the gene network is unfriendly to the setup of convolutional neural networks. To address these issues and build a robust classification model, we propose the Graph-Embedded Deep Feedforward Networks (GEDFN), to integrate external relational information of features into the deep neural network architecture. The method is able to achieve sparse connection between network layers to prevent overfitting. To validate the method's capability, we conducted both simulation experiments and a real data analysis using a breast cancer RNA-seq dataset from The Cancer Genome Atlas (TCGA). The resulting high classification accuracy and easily interpretable feature selection results suggest the method is a useful addition to the current classification models and feature selection procedures. The method is available at https://github.com/yunchuankong/NetworkNeuralNetwork. 
\keywords{classification, deep learning, gene expression, gene networks}
\end{abstract}

\section{Introduction}
\label{Introduction}
In recent years, more and more studies attempt to link clinical outcomes, such as cancer and other diseases, with gene expression or other types of profiling data. It is of great interest to develop new computational methods to predict disease outcomes based on profiling datasets that contain tens of thousands of variables. The major challenges in these data lie in the heterogeneity of the samples, and the sample size being much smaller than the number of predictors (genes), i.e. the $n<<p$ issue, as well as the complex correlation structure between the predictors. Thus the prediction task has been formulated as a classification problem combined with selection of predictors, solved by modern machine learning algorithms such as regression based methods \cite{liang2013sparse,algamal2015penalized}, support vector machines \cite{vanitha2015gene}, random forests \cite{kursa2014robustness,cai2015classification} and neural networks \cite{chen2014risk}. While these methods are aimed at achieving accurate classification performance, major efforts have also been put on selecting significant genes that effectively contribute to the prediction \cite{kursa2014robustness,cai2015classification}. However, feature selection is based on fitted predictive models and is conducted after parameter estimation, which causes the selection to rely on the classification methods rather than the structure of the feature space itself. Beside building robust predictive models, the feature selection also serves another important purpose - the functionality of the selected features (genes) can help unravel the underlying biological mechanisms of the disease outcome. 

Given the nature of the data, i.e. functionally associated genes tend to be statistically dependent and contribute to the biological outcome in a synergistic manner, a branch of gene expression classification research has been focused on integrating the relations between genes with classification methods, which helps in terms of both classification performance as well as learning the structure of feature space. A critical data source to achieve this goal has been gene networks. A gene network is a graph-structured dataset with genes as the graph vertices and their functional relations as graph edges. The functional relations are largely curated from existing biological knowledge \cite{pmid25632107,pmid25859942}. Each vertex in the network corresponds to a predictor in the classification model. Thus, it is expected that the gene network can provide useful information for a learning process where genes serve as predictors. Motivated by this fact, certain procedures have been developed where gene networks are employed to conduct feature selection prior to classification \cite{chuang2007network,wei2007incorporating}. Moreover, methods that integrate gene network information directly into classifiers have also been developed. For example, \cite{dutkowski2011protein} proposes the random forest-based method, where the feature sub-sampling is guided by graph search on gene networks when constructing decision trees. \cite{zhu2009network,lavi2012network} modify the objective function of the support vector machine with penalty terms defined according to pairwise distances between genes in the network. Similarly, \cite{kim2013network} develops logistic regression based classifier using regularization, where again a relational penalty term is introduced in the loss function. The authors of these methods have demonstrated that embedding expression data into gene network results in both better classification performance and more interpretable selected feature sets. 

With the clear evidence that gene networks can lead to novel variants of traditional classifiers, we are motivated to incorporate gene networks with deep feedforward networks (DFN), which is closely related to the state-of-the-art technique deep learning \cite{lecun2015deep}. Although nowadays deep learning has been constantly shown to be one of the most powerful tools in classification, its application in bioinformatics is limited \cite{min2016deep}. This is due to many reasons including the $n<<p$ issue, the large heterogeneity in cell populations and clinical subject populations, as well as inconsistent data characteristics across different laboratories, resulting in difficulties merging datasets. Consequently, the relatively small number of samples compared to the large number of features in a gene expression dataset obstructs the use of deep learning techniques, where the training process usually requires a large amount of samples such as in image classification \cite{ILSVRC15}. Therefore, there is a need to modify deep learning models for disease outcome classification using gene expression data, which naturally leads us to the development a variant of deep learning models specifically fitting the practical situation with the help of gene networks. 

Incorporating gene networks as relational information in the feature space into DFN classifiers is a natural option to achieve sparse learning with less parameters compared to usual DFN. However, to the best of our knowledge, few existing work has been done on this track. \cite{bruna2013spectral,henaff2015deep} started the direction of sparse deep neural networks for graph-structured data. The authors developed hierarchical locally connected network architectures with newly defined convolution operations on graph-structured data. The methods have novel mathematical formulation, however, the applications are yet to be generalized. In both of the two papers, by using the two benchmark datasets MINST \cite{lecun-mnisthandwrittendigit-2010} and ImageNet \cite{ILSVRC15} respectively, the authors have treated 2-D grid images as a special form of graph-structured data in their experiments. This is based on the fact that a image can be regarded as a graph in which each pixel is a vertex connected with four neighbors in the four directions. However, graph-structured data can be much more complex in general, as the degree of each vertex can vary widely, and the edges do not have orientations as in image data. For a gene network, the degree of vertices is power-law distributed as the network is scale-free \cite{Kolaczyk:2009:SAN:1593430}. In this case, convolution operations are not easy to define. In addition, with tens of thousands of vertices in the graph, applying multiple convolution operations results in huge number of parameters, which easily leads to over-fitting given the small number of training samples. By taking an alternative approach of modifying a usual DFN, our newly proposed graph-embedded DFN can serve as a convenient tool to fill the gap. It avoids over-fitting in the $n<<p$ scenario, as well as achieves good feature selection results using the capabilities of DFN. 

The paper is organized as follows: Section \ref{Methods} reviews usual deep feedforward networks and illustrates our network-embedded architecture. Section \ref{se} compares the performance of our method with two related approaches using synthetic datasets, followed by the real application of a breast cancer dataset in Section \ref{rda}. Finally, conclusions and discussion are presented in Section \ref{Conclusion}. 

\section{Methods}
\label{Methods}
\subsection{Deep feedforward networks}
\label{dfn}
A deep feedforward network (DFN, or \textit{deep neural network} (DNN), \textit{multilayer perceptron} (MLP)) with $l$ hidden layers has a standard architecture 
\begin{align*}
Pr(\mathbf{y}|\mathbf{X},\mathbf{\theta})&=f(\mathbf{Z}_{out}\mathbf{W}_{out}+\mathbf{b}_{out}) \\
	\mathbf{Z}_{out}&=\sigma(\mathbf{Z}_{l}\mathbf{W}_{l}+\mathbf{b}_l) \\
	\dots \\
	\mathbf{Z}_{k+1}&=\sigma(\mathbf{Z}_{k}\mathbf{W}_{k}+\mathbf{b}_k) \\
	\dots \\
	\mathbf{Z}_{1}&=\sigma(\mathbf{X}\mathbf{W}_{in}+\mathbf{b}_{in}), 
\end{align*}
where $\mathbf{X}\in \mathcal{R}^{n\times p}$ is the input data matrix with $n$ samples and $p$ features, $\mathbf{y}\in \mathcal{R}^n$ is the outcome vector containing classification labels, $\mathbf{\theta}$ denotes all the parameters in the model, $\mathbf{Z}_{out}$ and $\mathbf{Z}_{k}, k=1,\dots,l-1$ are hidden neurons with corresponding weight matrices $\mathbf{W}_{out}$, $\mathbf{W}_{k}$ and bias vectors $\mathbf{b}_{out}$, $\mathbf{b}_{k}$. The dimensions of $\mathbf{Z}$ and $\mathbf{W}$ depend on the number of hidden neurons $h_{in}$ and $h_k, k=1,\dots,l$, as well as the input dimension $p$ and the number of classes $h_{out}$ for classification problems. In this paper, we mainly focus on binary classification problems hence the elements of $\mathbf{y}$ simply take binary values and $h_{out}\equiv 2$. $\sigma(\cdot)$ is the activation function such as sigmoid, hyperbolic tangent or rectifiers. $f(\cdot)$ is the softmax function converting values of the output layer into probability prediction i.e. 
\begin{equation*}
	p_{i}=f(\mu_{i1})=\frac{e^{\mu_{i1}}}{e^{\mu_{i0}}+e^{\mu_{i1}}}
\end{equation*}
where
\begin{align*}
    p_{i}:&=Pr(y_i=1|\mathbf{x}_i)\\
    \mu_{i0}:&=[\mathbf{z}_{i}^{(out)}]^T\mathbf{w}_{0}^{(out)}+\mathbf{b}_{i}^{(out)}\\
    \mu_{i1}:&=[\mathbf{z}_{i}^{(out)}]^T\mathbf{w}_{1}^{(out)}+\mathbf{b}_{i}^{(out)}, 
\end{align*} for binary classification where $i=1,\dots,n$. 

The parameters to be estimated in this model are all the weights and biases. For a training dataset given true labels, the model is trained using a stochastic gradient decent (SGD) based algorithm \cite{goodfellow2016deep} by minimizing the cross-entropy loss function 
\begin{equation*}\label{loss function}
	\mathcal{L}(\mathbf{\theta})=-\frac{1}{n}\sum_{i=1}^{n}\{y_i log(\hat{p}_i)+(1-y_i)log(1-\hat{p}_i)\},
\end{equation*}
where again $\mathbf{\theta}$ denotes all the model parameters, and $\hat{p}_i$ is the fitted value of $p_i$. More details about DFN can be found in \cite{goodfellow2016deep}.

\subsection{Graph-embedded deep feedforward networks}
\label{gedfn}
Our newly proposed DNN model is based on two main assumptions. The first assumption is that neighboring features on a known scale-free feature network or feature graph\footnote{Since in this paper we interchangeably discuss feature networks and neural networks, to avoid confusion, the equivalent term ``graph" is used to refer to the feature network from now on, while ``networks" naturally refer to neural networks.} tend to be statistically dependent. The second assumption is that only a small number of features are true predictors for the outcome, and the true predictors tend to form cliques in the feature graph. These assumptions have been commonly used and justified in previous works reviewed in Section \ref{Introduction}.  

To incorporate the known feature graph information to DNN, we propose the graph-embedded deep feedforward network (GEDFN) model. The key idea is that, instead of letting the input layer and the first hidden layers to be fully connected, we embed the feature graph in the first hidden layer so that a fixed informative sparse connection can be achieved. 

Let $\mathbf{G}=(V, E)$ be a known graph of $p$ features, with $V$ the collection of $p$ vertices and $E$ the collection of all edges connecting vertices. A common representation of a graph is the corresponding adjacency matrix $A$. Given a graph $\mathbf{G}$ with $p$ vertices, the adjacency $A$ is a $p\times p$ matrix with 
\begin{equation*}
A_{ij} = 
	\begin{cases}
		1, & \text{if}\ V_i\ \text{and}\ V_j\ \text{are connected},\ \forall i,j = 1,\dots,p \\
      	0, & \text{otherwise}.
	\end{cases}
\end{equation*}
In our case $A$ is symmetric since the graph is undirected. Also, we require $A_{ii}=1$ meaning each vertex is regarded to connecting itself. 

Now to mathematically formulate our idea, we construct the DNN such that the dimension of the first hidden layer ($h_{in}$) is the same as the original input i.e. $h_{in}=p$, hence $\mathbf{W}_{in}$ has a dimension of $p\times p$. Between the input layer $\mathbf{X}$ and the first hidden layer $\mathbf{Z}_1$, instead of fully connecting the two layers with $
\mathbf{Z}_{1}=\sigma(\mathbf{X}\mathbf{W}_{in})+\mathbf{b}_{in}$, we have 
\begin{equation*}
	\mathbf{Z}_{1}=\sigma(\mathbf{X}(\mathbf{W}_{in}\odot A)+\mathbf{b}_{in})
\end{equation*}
where the operation $\odot$ is the Hadamard (element-wise) product. Thus, the connections between the first two layers of the feedforward network are ``filtered" by the feature graph adjacency matrix. Through the one-to-one $\mathcal{R}: p\to p$ transformation, all features have their corresponding hidden neurons in the first hidden layer. A feature can only feed information to hidden neurons that correspond to features connecting to it in the feature graph. 

Specifically, let $\mathbf{x}_i=(x_{i1},\dots,x_{ip})^T, i=1,\dots,n$ be any instance (one row) of the input matrix $\mathbf{X}$, in usual DFN, the first hidden layer of this instance is calculated as 
\begin{equation*}	\mathbf{z}_{i}^{(1)}=\sigma([\sum_{j=1}^{p}x_{ij}w_{1j}^{(in)}+b_1^{(in)},\dots,\sum_{j=1}^{p}x_{ij}w_{h_{in}j}^{(in)}+b_{h_{in}}^{(in)}]^T),
\end{equation*}
where $\mathbf{z}_{i}^{(1)}$ is the $i$-th row of $\mathbf{Z}_1$, and $w_{kj}^{(in)}, b_k^{(in)}, k=1,\dots,h_{in}$ are the weight and bias for this layer. Now in our model, $h_{in}=p$ and each $w_{kj}^{(in)}$ is multiplied by an indicator function i.e.
\begin{equation*}
	\mathbf{z}_{i}^{(1)}=\sigma([\sum_{j=1}^{p}x_{ij}w_{1j}^{(in)}\mathcal{I}		(A_{1j}=1)+b_1^{(in)},\dots,\sum_{j=1}^{p}x_{ij}w_{pj}^{(in)}\mathcal{I}(A_{pj}=1)+b_p^{(in)}]^T).
\end{equation*}
Therefore, the feature graph helps achieve sparsity for the connection between the input layer and the first hidden layer. 

\subsection{Detailed model settings}
For the choice of activation functions in DNN, the rectified linear unit (ReLU) \cite{nair2010rectified} with the form (in scalar case)
\begin{equation*}
	\sigma_{ReLU}(x)=max(x,0)
\end{equation*} is employed. This activation has an advantage over sigmoid and hyperbolic tangent as it can avoid the vanishing gradient problem \cite{hochreiter2001gradient} during training using SGD. To train the DNN model, we choose the Adam optimizer \cite{DBLP:journals/corr/KingmaB14}, which is the most widely used variant of traditional gradient descent algorithms in deep learning. Also, we use the mini-batch training strategy by which the optimizer randomly trains a small proportion of the samples in each iteration. Details about the Adam optimizer and mini-batch training can be found in \cite{goodfellow2016deep,DBLP:journals/corr/KingmaB14}.

\begin{figure}
\centering
	\includegraphics[scale=0.5]{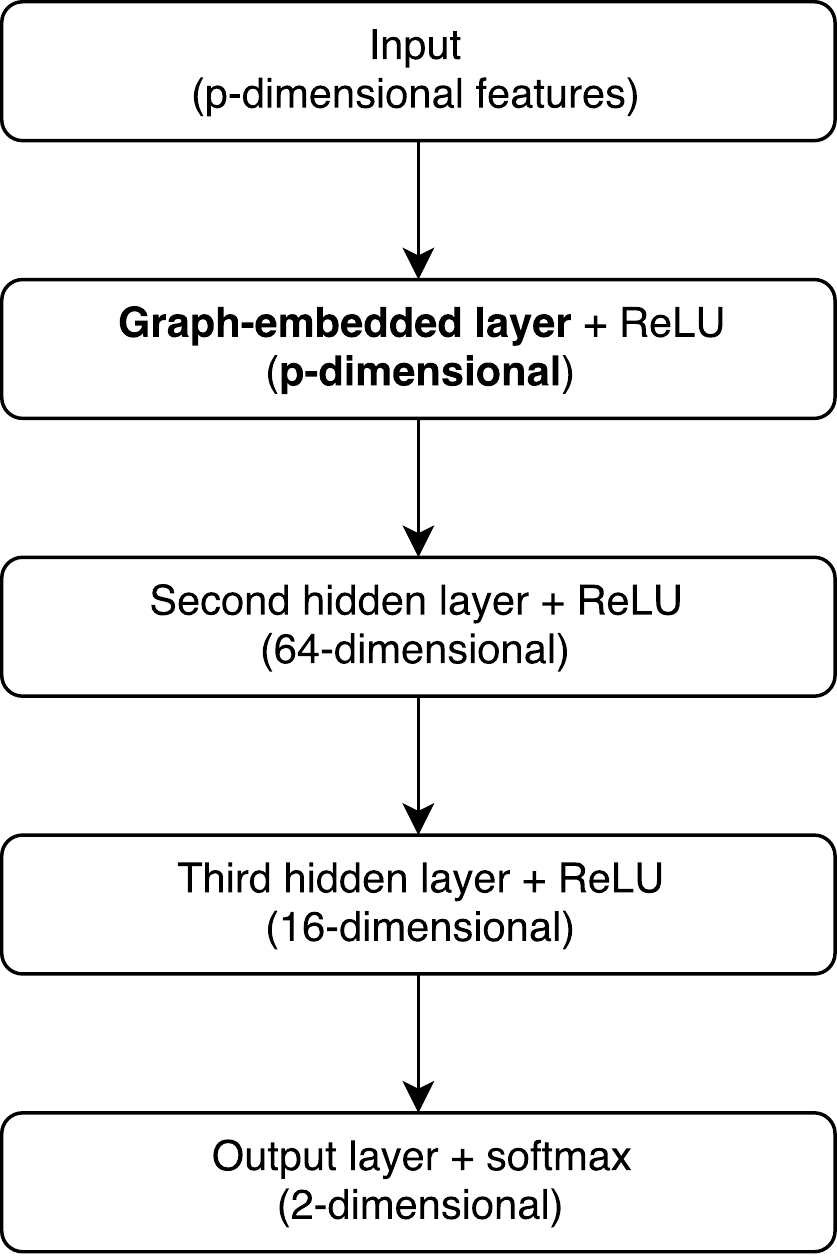}\\
\caption{Network architecture of the GEDFN model for experiments in Section \ref{se} and Section \ref{rda}.}\label{architecture}
\end{figure}

The classification performance of a DNN model is associated with many hyper-parameters, including architecture related parameters such as the number of layers and the number of hidden neurons in each layer, regularization related parameters such as the dropout proportion and the penalty scale of regularizers, model training related parameters such as the learning rate and the batch size. These hyper-parameters can be fine tuned using advanced hyper-parameter training algorithm such as Bayesian Optimization \cite{mockus2012bayesian}, however, as the hyper-parameters are not of primary interest in our work, in later sections, we simply tune them using grid search in a feasible hyper-parameter space. A visualization of our fine tuned GEDFN model for simulation and real data experiments is shown in Fig. \ref{architecture}.   
  
\section{Simulation Experiments}
\label{se}
We conduct extensive simulation experiments based on our assumptions of data. The goal of the experiments is to mimic disease outcome classification using gene expression and network data, and explore the performance of our new method compared to the usual DFN. Robustness is also tested as we simulate datasets that do not fully satisfy the main assumptions. The method is applied to examine whether it can still achieve a reasonable performance (i.e. at least as good as a usual DFN). 

\subsection{Synthetic data generation}
\label{sdg}
For a given number of features $p$, we employ the preferential attachment algorithm proposed by \cite{barabasi1999emergence} to generate a scale-free feature graph. The $p\times p$ distance matrix $D$ recording pairwise distances among all vertices is then calculated. Next, we transform the distance matrix into a covariance matrix $\Sigma$ by letting 
\begin{equation*}
	\Sigma_{ij}=0.7^{D_{ij}}, i,j=1,\dots,p.
\end{equation*} Here by convention the diagonal elements of $D$ are all zeros meaning the distance between a vertex to itself is zero. 

After simulating the feature graph and obtaining the covariance matrix of features, we generate $n$ multivariate Gaussian samples as the input matrix $\mathbf{X}=(\mathbf{x}_1,\dots,\mathbf{x}_n)^T$ i.e.
\begin{equation*}
	\mathbf{x}_i\sim \mathcal{N}(\mathbf{0},\Sigma), i=1,\dots\,n,
\end{equation*}
where $n\ll p$ for imitating gene expression data. To generate outcome variables, we first select a subset of features to be the ``true" predictors. Following our assumptions mentioned in Section \ref{gedfn}, we intend to select cliques in the feature graph. Among vertices with relatively high degrees, part of them are randomly selected as ``cores", and part of the neighboring vertices of cores are also selected. Denoting the number of true predictors as $p_0$, we sample a set of parameters $\mathbf{\beta}=(\beta _1,\dots,\beta _{p_0})^T$ and an intercept $\beta_0$ within a certain range. In our experiments, we first sample $\beta$'s from $(0.1, 0.2)$, so that the signal will neither be too strong nor too weak. Also, some of the parameters are randomly turned into negative, so that we accommodate both positive and negative coefficients. Finally, the outcome variable $\mathbf{y}$ is generated through a logistic regression model
\begin{align*}
	Pr(y_i=1|\mathbf{x_i})&=logit^{-1}(\mathbf{x_i}^T\mathbf{\beta}+\beta_0)\\
    y_i&=\mathcal{I}(Pr(y_i=1|\mathbf{x_i})>t),i=1,\dots\,n,
\end{align*}
where $t$ is a threshold and $logit(\cdot)$ is the logit function
\begin{equation*}
	logit(x) = log(\frac{x}{1-x}).
\end{equation*} The inverse $logit^{-1}$ is equivalent to a binary class softmax function. 

Following the above procedure, we simulate a set of synthetic datasets with 5,000 features and 400 samples. We compare our method with the usual DFN and another feature graph-embedded classification method network-guided forest (NGF) \cite{dutkowski2011protein} mentioned in Section \ref{Introduction}. In gene expression data, the number of true predictors account for only a small proportion of the features, i.e. the signal-to-noise ratio is extremely low. Taking this aspect into consideration, we examine different numbers, i.e. 40, 80, 120 160, and 200, of true predictors, corresponding to 2, 4, 6, 8 and 10 cores among all the high-degree vertices in the feature graph. However, in reality, the true predictors may not be perfectly distributed in the feature graph as cliques. Instead, some of the true predictors, which we call ``singletons", can be quite scattered. To create this possible circumstance, we simulate three series of datasets with singleton proportions 0\%, 50\% and 100\% among all the true predictors. Therefore, we investigate three situations where all true predictors are in cliques, half of the true predictors are singletons, and all of the true predictors are scattered in the graph, respectively.     

\subsection{Simulation results}
\label{sr}
In our simulation studies, as shown in Fig. \ref{architecture}, the GEDFN had three hidden layers, where the first hidden layer was the graph adjacency embedded layer. Thus the dimension of its output is the same as the input, namely 5,000. The second and third hidden layers had 64 and 16 hidden neurons respectively, which are the same for the usual DFN. The number of the first layer hidden neurons in the usual DFN was tuned using grid search. When the number of true predictors was 40, the first hidden layer had 512 neurons; as we increased the number of true predictors, 1,024 hidden neurons achieved slightly better performance. For each of the data generation settings, 10 datasets were generated, and the GEDFN, DFN, and NGF methods were applied on the data. For each simulated dataset, we randomly split the dataset into training and testing sets at a 4:1 ratio. The final testing set classification accuracy results were then averaged across the ten datasets. 
All the classification results were evaluated by the area under the receiver operating characteristic (ROC) curve (AUC).   

\begin{figure}
\centering
	\begin{minipage}[b]{0.32\textwidth}
		\includegraphics[width=\textwidth]{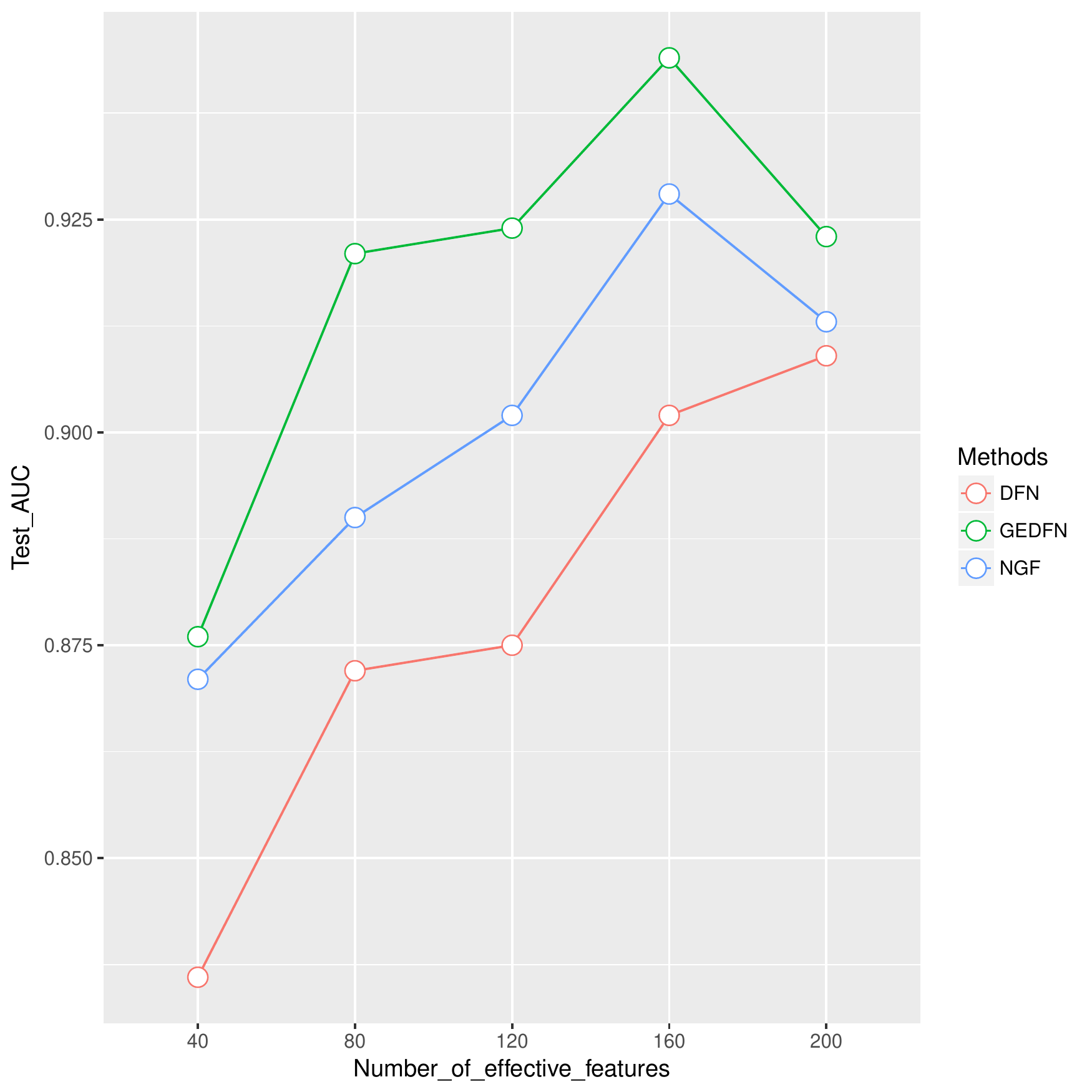}\\
        \centering{(a)}
	\end{minipage}
    \begin{minipage}[b]{0.32\textwidth}
		\includegraphics[width=\textwidth]{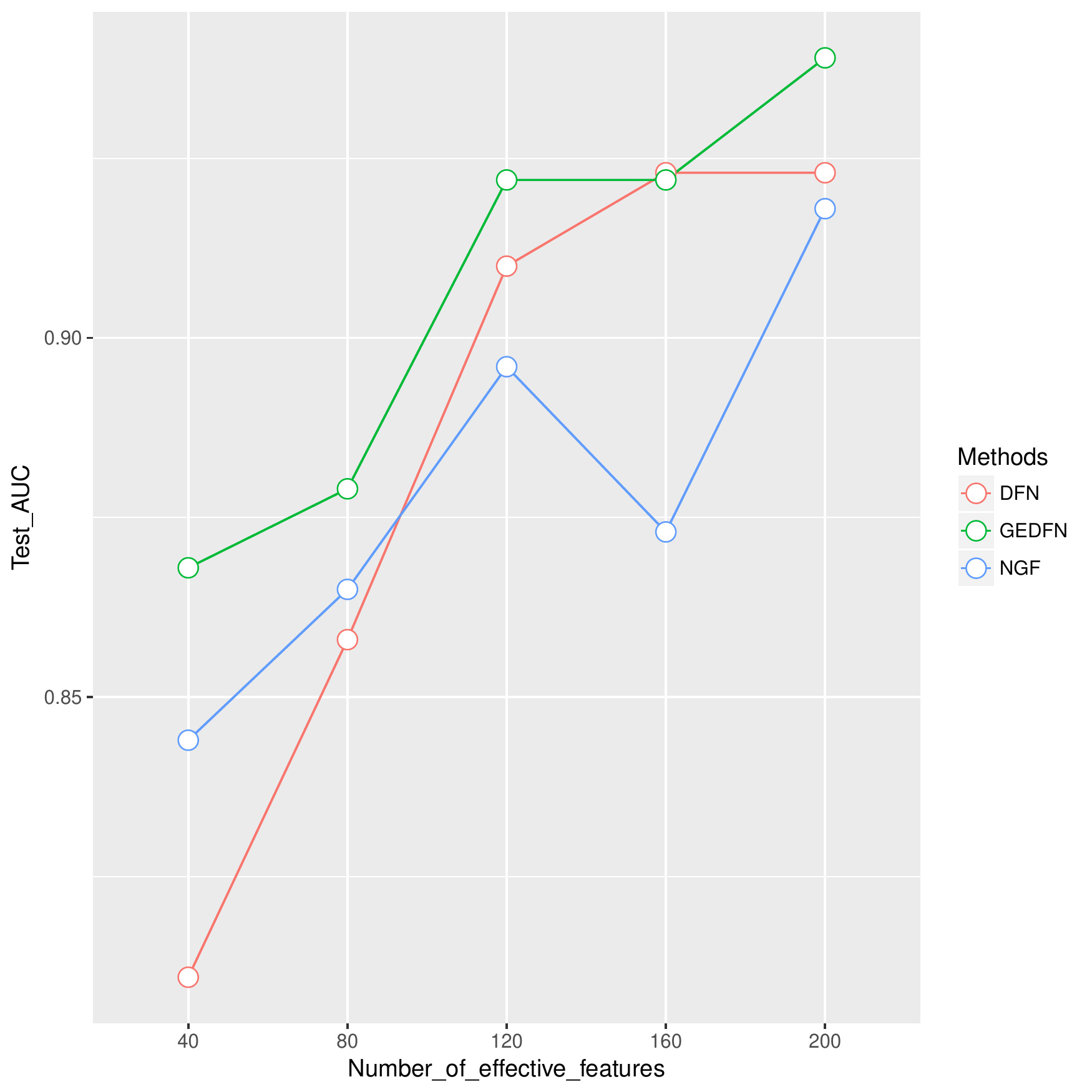}\\
        \centering{(b)}
	\end{minipage}
    \begin{minipage}[b]{0.32\textwidth}
		\includegraphics[width=\textwidth]{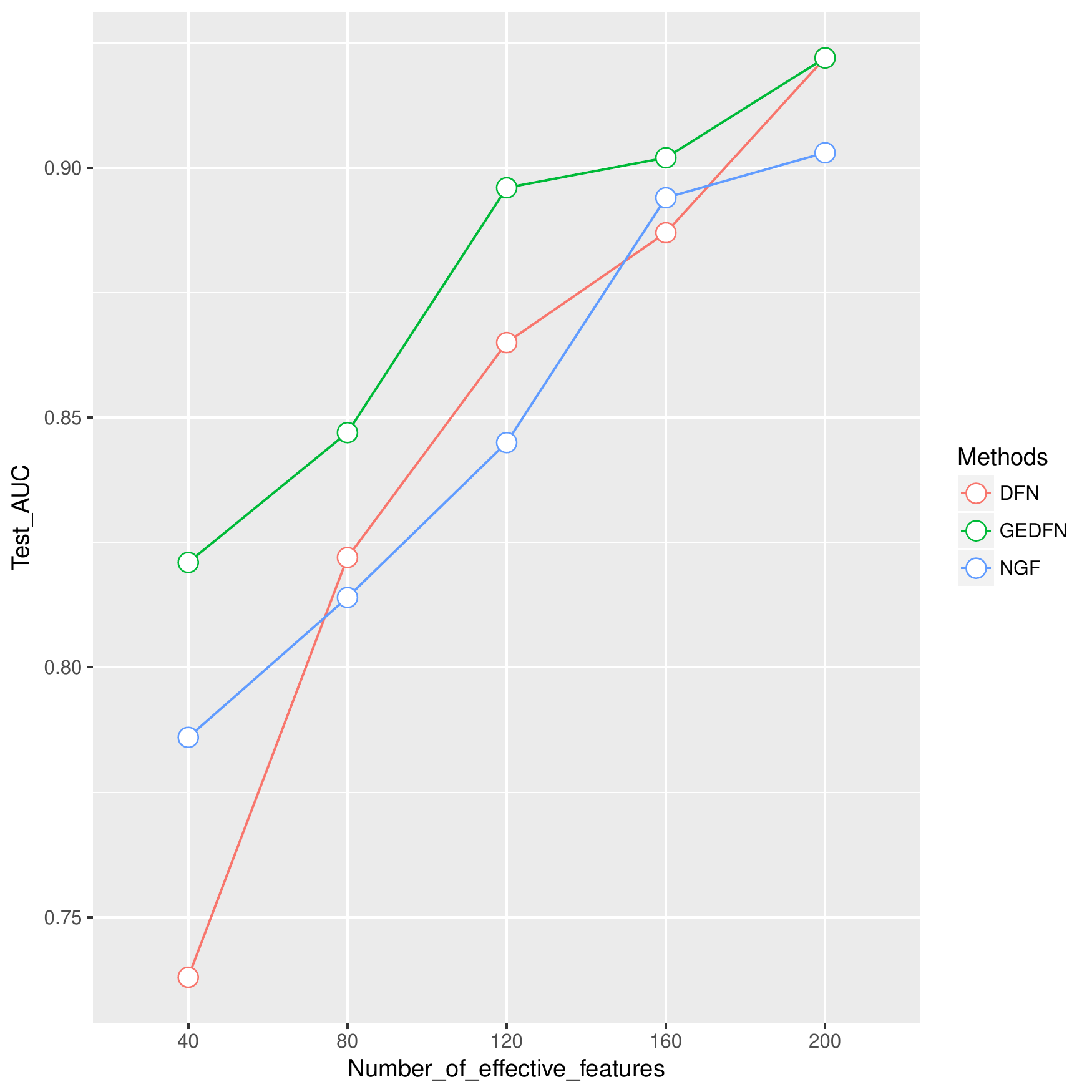}\\
        \centering{(c)}
	\end{minipage}
\caption{Plots of the classification comparison in Table \ref{simulation_table}. Singleton proportions: (a) 0\%, (b) 50\%, (c) 100\%. }\label{simulation_figures}
\end{figure}

Table \ref{simulation_table} and Fig. \ref{simulation_figures} shows the results of the simulation experiments. Corresponding to the case that singleton proportion is 0\%, Fig. \ref{simulation_figures}(a) shows the two feature graph integrated methods outperformed DFN with no feature graph information. As the number of true predictors increased, all of the methods performed better as there were more signals in the feature set. As the singleton proportion increased to 0.5 (Fig. \ref{simulation_figures}(b)), GEDFN was still the best among the three while the performance of NGF decreased. In Fig. \ref{simulation_figures}(c), the difference between GEDFN and DFN is not very obvious, since there was essentially no feature graph information with singleton proportion at 100\%. At the same time, NGF performed worse than neural network methods. It is also noted that with the increase of singleton proportions, the performance of DFN became worse as well, which is because neural network methods inherently tackle correlated features, and the correlation decreased as the number of singletons increased. Therefore, although not as directly affected as GEDFN and NGF, increased singleton proportions also deteriorated the performance of DFN.

\begin{table}
\centering
\small 
\caption{Classification comparison of the graph-embedded deep feedforward network (GEDFN) method, the network-guided forest (NGF) method and the usual DFN. Statistics are the classification accuracies measured by AUC.} 

\begin{tabular}{c|ccccc|ccccc|ccccc}
\hline 
\% singletons & \multicolumn{5}{c|}{0\%} & \multicolumn{5}{c|}{50\%} & \multicolumn{5}{c}{100\%}\tabularnewline
\hline 
\# true predictors & 40 & 80 & 120 & 160 & 200 & 40 & 80 & 120 & 160 & 200 & 40 & 80 & 120 & 160 & 200\tabularnewline
\hline 
DFN & 0.836 & 0.872 & 0.875 & 0.902 & 0.909 & 0.811 & 0.858 & 0.91 & 0.923 & 0.923 & 0.738 & 0.822 & 0.865 & 0.887 & 0.922\tabularnewline
\hline 
GEDFN & 0.876 & 0.921 & 0.924 & 0.944 & 0.923 & 0.868 & 0.879 & 0.922 & 0.922 & 0.939 & 0.821 & 0.847 & 0.896 & 0.902 & 0.922\tabularnewline
\hline 
NGF & 0.871 & 0.89 & 0.902 & 0.928 & 0.913 & 0.844 & 0.865 & 0.896 & 0.873 & 0.918 & 0.786 & 0.814 & 0.845 & 0.894 & 0.903\tabularnewline
\hline 
\end{tabular}

\label{simulation_table}
\end{table}

In summary, the simulation experiments demonstrated that compared to the NGF method and the usual DFN method, our newly proposed GEDFN model had better classification accuracies in cases where true predictors were concentrated on cliques in the feature network. When the number of singletons increased, the feature network could hardly provide any signal and the performance of GEDFN declined to the same level of the usual DFN model. 

\section{Real data application}
\label{rda}
\subsection{Datasets}

We applied our GEDFN method to the Cancer Genome Atlas (TCGA) breast cancer (BRCA) RNA-seq dataset \cite{koboldt2012comprehensive}. The dataset consists of a gene expression matrix with 20,532 genes of 707 cancer patients, as well as the clinical data containing various disease status measurements. The gene network came from the HINT database \cite{pmid22846459}. In this proof-of-concept study, we were interested in the relation between gene expression and a molecular subtype of breast cancer - the tumor's Estrogen Receptor (ER) status. ER is expressed in more than 2/3 of breast tumors, and plays a critical role in tumor growth \cite{pmid12829800}.

After screening genes that were not involved in the gene network, a total of 9,211 genes were used as the final feature set in our classification. For each gene, the expression value was Z-score transformed i.e. the expression value minus the mean across all patients and then divided by the standard error. 

\subsection{Model fitting and evaluation of feature importance}

In the analysis of gene expression data, not only are we concerned about the classification result, but also it is of great interest to find features that significantly contribute to the classification, as they can reveal the underlying biological mechanisms. In our example, besides training a well-performed classification model, we also intended to figure out which genes play important roles in prediction. After fitting the GEDFN model, we employed the partial derivative method proposed in \cite{dimopoulos1995use} to calculate an approximate score for each gene as the variable importance.

The main idea of the partial derivative method is that the importance of a specific variable is reflected by its impact on the change of prediction when its value changes. The ratio between change in prediction and change in the variable is thus the partial derivative of the predicted probability with respect to the variable i.e. $\frac{\partial p_i}{\partial x_{ij}}, i=1,\dots,n, j=1,\dots,p$. To obtain the partial derivative, one straightforward approximation is to compute it numerically. The procedure is as the following: after fitting the model using training data and obtaining the testing data prediction, for each gene in the testing dataset, we increase a small proportion $\delta$ (say, 5\%) of its expression value in all the testing samples while fixing other gene values unchanged. Re-running the model using this perturbed testing set, the new prediction is then obtained for each sample. Next, the impact of changing the gene value is computed as the difference between the new prediction and the original prediction. Finally, the average ratio between the prediction difference and the value changed in expression across samples can serves as the numerical partial derivative. Equation \ref{importance_score} shows a mathematical expression for this numerical partial derivative $s_j$ for gene $j$:  
\begin{equation}\label{importance_score}
s_j=\frac{1}{n}\sum_{i=1}^{n}\frac{\tilde{p}_{ij}-\hat{p}_{i}}{\mid \delta x_{ij}\mid}, j=1,\dots,p,
\end{equation}
where $\tilde{p}_{ij}$ and $\hat{p}_{i}$ are the perturbed prediction and the original prediction respectively, and $\delta$ is the small percentage of the change in the gene expression value. This calculation procedure is repeated for every gene. To compare the effect size of genes, we use the absolute value $|s_j|$ as the importance score. A ranked importance list is then obtained by sorting the genes according to their scores.

\subsection{Results}

\begin{figure}[H]
\centering
	\begin{minipage}[b]{0.45\textwidth}
		\includegraphics[width=\textwidth]{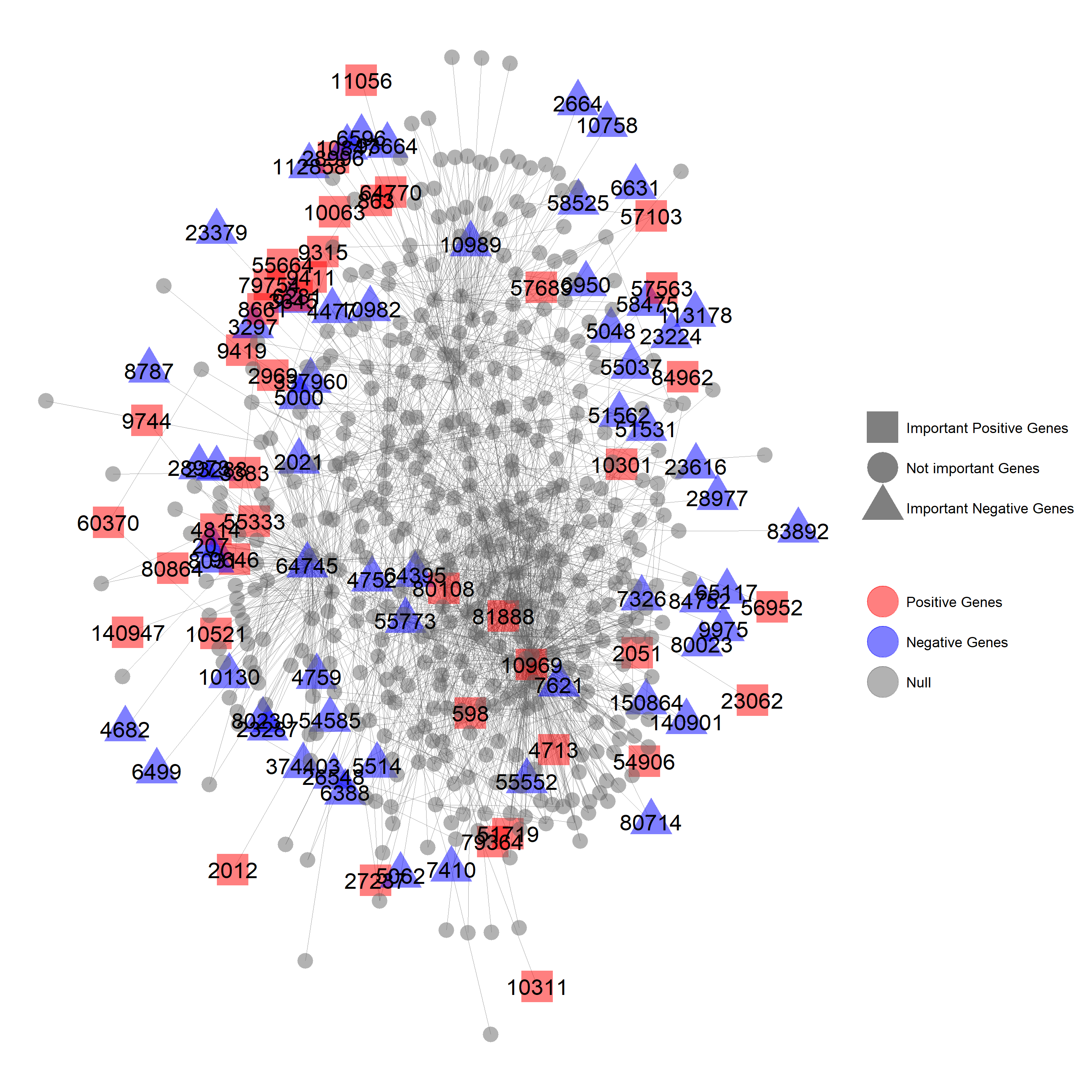}\\
        \centering{(a)}
	\end{minipage}
    \begin{minipage}[b]{0.45\textwidth}
		\includegraphics[width=\textwidth]{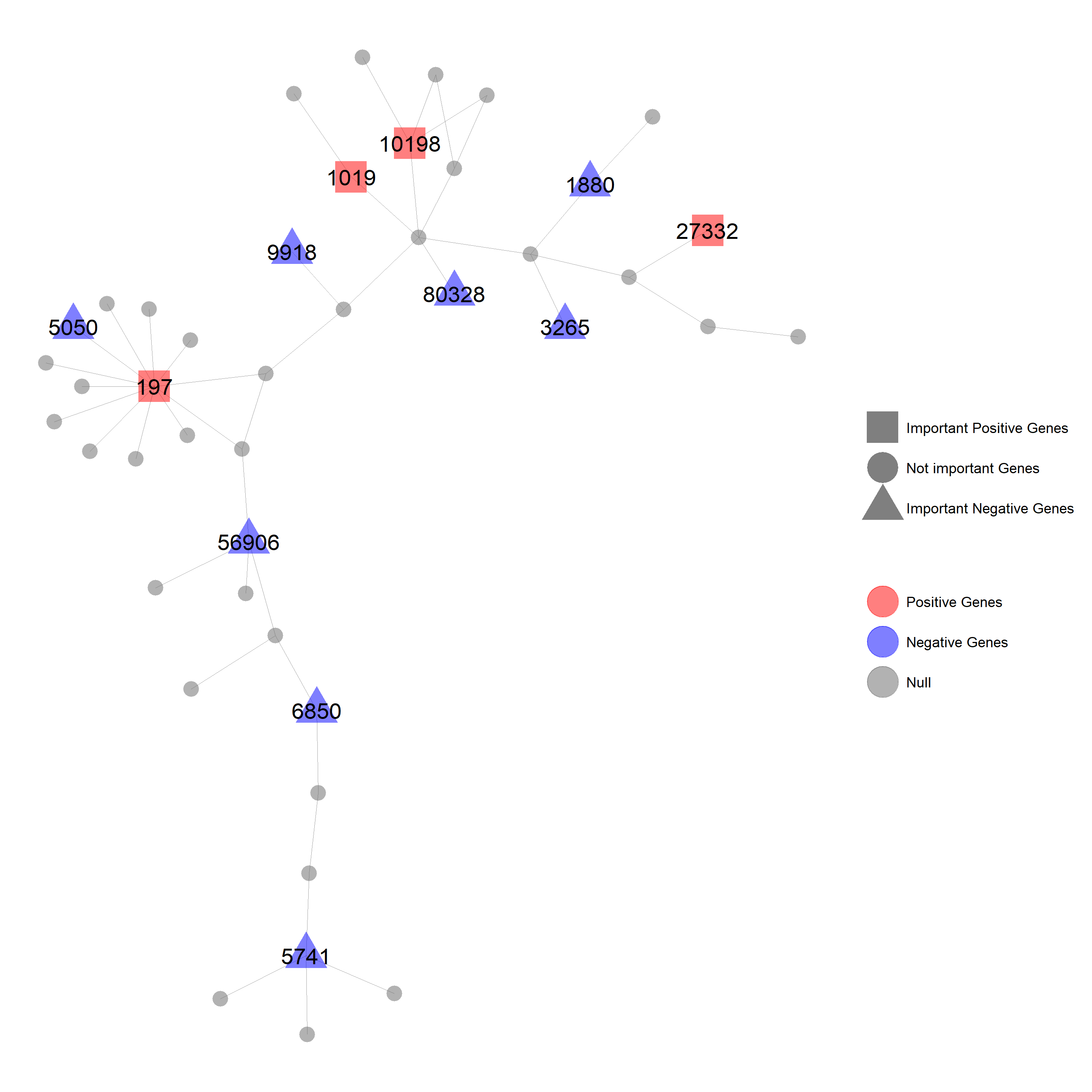}\\
        \centering{(b)}
	\end{minipage}
\caption{Example modules from the top 5\% ranked gene list. }\label{annotation_figures}
\end{figure}

Using the HINT network architecture as in Section \ref{sr}, we tested our GEDFN method on the BRCA data with ten repeated experiments. The average testing AUC is 0.938, indicating that the method is well applicable for gene expression-based  classification problem. The ranked effective genes list was also obtained by averaging gene importance scores across the 10 repeated analysis. We conducted further functional analyses for the top 5\% ranked genes to interpret the feature importance of our model from the biological point of view. We conducted community detection on the network containing the top-ranked genes and their one-step neighbors \cite{clauset2004finding}. Edges were weighted such that edges between top-ranked genes received a weight of 15, while other edges received a weight of 1. Sixteen modules were selected, and the corresponding top gene groups were tested for functionality through GO enrichment analysis \cite{pmid17098774}. Fig. \ref{annotation_figures} shows two examples of the 16 modules. 

The largest module shown in Fig. \ref{annotation_figures}a consists of 34 positive genes, i.e. the increased expression of which increases the probability of the sample being classified into the ``ER+" group, and 27 negative genes, i.e. the increased expression of which decreases the probability of the sample being classified as ``ER+". Functional analysis using GOstats reveals several key biological processes overrepresented by the top ranked genes in this module. The most significant biological process is regulation of execution phase of apoptosis. ER is known to be closely related to the regulation of the apoptosis process. In the presence of estrogen, chemotherapy-induced apoptosis is suppressed in ER+ breast cancer cells \cite{pmid23077249}. It has been shown that siRNA-mediated suppression of protein tyrosine phosphatase $\alpha$ (PTP$\alpha$) induces apoptosis in ER- but not in ER+ breast cancer cells \cite{pmid18183590}. 

The second biological process found was centrosome localization. Centrosome amplification mediated be estrogen leads to chromosomal instability, which is a critical process in breast oncogenesis \cite{pmid15601761,pmid25499220}. The results indicate the lack of estrogen receptor in the ER- group is associated with a distinct gene expression patter in centrosome localization genes.

As shown by the third and fourth biological processes, a number of regulators of phosphorylation, many of which are involved in mitogen-activated  protein  kinase  (MAPK)  cascade, were associated with ER status. The MAPK cascade plays an integral role in estrogen signaling. Ras-MAPK cascade modulates the activity of ER \cite{pmid7491495}, and MAPK can be activated by estradiol (E2) independent of transcriptional changes \cite{pmid10200323}. 

\begin{table}[H]
\centering
\normalsize 
\caption{GO biological process enrichment analysis of the top (colored) genes in Fig. \ref{annotation_figures}a. The top 5 terms are shown after manually removing redundant terms.
}
\begin{tabular}{llll}
\hline 
\textbf{GOBPID} & \textbf{Term} & \textbf{Pvalue} & \textbf{Genes in the GO Term}\tabularnewline
GO:1900117 & regulation of execution phase of apoptosis & 1.7E-05 &  598;2021;3297;5062\tabularnewline
\hline 
GO:0051642 & centrosome localization & 0.00042 &  4682;5048;23224\tabularnewline
\hline 
GO:0042326 & negative regulation of phosphorylation & 0.00049 &  207;863;5048;5062;8661;\tabularnewline
 &  &  &  51562;55333;57103;\tabularnewline
 &  &  &   57689;84962;374403\tabularnewline
\hline 
GO:0043409 & negative regulation of MAPK cascade & 0.0015 &  207;5048;8661;51562;\tabularnewline
 &  &  &   55333;374403\tabularnewline
\hline 
GO:0043087 & regulation of GTPase activity & 0.0020 & 2664;3383;5048;6281;\tabularnewline
 &  &  &   7410;8787;9411;9744;\tabularnewline
 &  &  &   23616;27237;84962;374403\tabularnewline
\hline 
\end{tabular}

\label{annotation_table10}
\end{table}

A number of genes involved in GTPase activity regulation are also transcriptionally associated with ER status. Through the G protein-coupled receptor GPR30, estrogen triggers the activation of the MAPKs Erk-1 and Erk-2 \cite{pmid11773440}. At the same time, activation of GPR30 by the receptor-specific agonist G-1 induces mitochondrial-related apoptosis \cite{pmid25275589}.

The module in Fig. \ref{annotation_figures}b is much smaller. Two of the positive genes (197, 1019) are involved in the insulin receptor signaling pathway, which has been shown to cross-talk with the estrogen signaling pathway in estrogen-dependent breast cancer \cite{pmid18991569}. Five of the selected genes (1019;1880;3265;5741;6850) belong to the process ``positive regulation of cell proliferation", three of which (1880;3265;6850) are part of the ERK1 and ERK2 cascade, which as we discussed before, is triggered by estrogen through GPR30 \cite{pmid11773440}. 

Overall, the real data results based on RNA-seq data and HINT database clearly demonstrated the method can achieve good classification performance by selecting biologically relevant genes. 

\section{Conclusion}
\label{Conclusion}
We presented a new deep feedforward network classifier embedding feature graph information. It achieves sparse connected neural networks by constraining connections between the input layer and the first hidden layer according to the feature graph. Simulation experiments have shown its relatively higher classification accuracy compared to existing methods, and the real data application demonstrated the utility of the new model. 

\section*{Acknowledgements}
\label{Acknowledgements}
This study was partially funded by National Institutes of Health [grant number R01GM124061]. The authors thank Dr. Hao Wu and Dr. Jian Kang for helpful discussions.

\bibliographystyle{splncs03}
\bibliography{main} 

\begin{thebibliography}{10}
\providecommand{\url}[1]{\texttt{#1}}
\providecommand{\urlprefix}{URL }

\bibitem{algamal2015penalized}
Algamal, Z.Y., Lee, M.H.: Penalized logistic regression with the adaptive lasso
  for gene selection in high-dimensional cancer classification. Expert Systems
  with Applications  42(23),  9326--9332 (2015)

\bibitem{pmid23077249}
Bailey, S.T., Shin, H., Westerling, T., Liu, X.S., Brown, M.: {{E}strogen
  receptor prevents p53-dependent apoptosis in breast cancer}. Proc. Natl.
  Acad. Sci. U.S.A.  109(44),  18060--18065 (Oct 2012)

\bibitem{barabasi1999emergence}
Barab{\'a}si, A.L., Albert, R.: Emergence of scaling in random networks.
  science  286(5439),  509--512 (1999)

\bibitem{bruna2013spectral}
Bruna, J., Zaremba, W., Szlam, A., LeCun, Y.: Spectral networks and locally
  connected networks on graphs. arXiv preprint arXiv:1312.6203  (2013)

\bibitem{cai2015classification}
Cai, Z., Xu, D., Zhang, Q., Zhang, J., Ngai, S.M., Shao, J.: Classification of
  lung cancer using ensemble-based feature selection and machine learning
  methods. Molecular BioSystems  11(3),  791--800 (2015)

\bibitem{chen2014risk}
Chen, Y.C., Ke, W.C., Chiu, H.W.: Risk classification of cancer survival using
  ann with gene expression data from multiple laboratories. Computers in
  biology and medicine  48,  1--7 (2014)

\bibitem{pmid25632107}
Chowdhury, S., Sarkar, R.R.: {{C}omparison of human cell signaling pathway
  databases--evolution, drawbacks and challenges}. Database (Oxford)  2015
  (2015)

\bibitem{chuang2007network}
Chuang, H.Y., Lee, E., Liu, Y.T., Lee, D., Ideker, T.: Network-based
  classification of breast cancer metastasis. Molecular systems biology  3(1),
  140 (2007)

\bibitem{clauset2004finding}
Clauset, A., Newman, M.E., Moore, C.: Finding community structure in very large
  networks. Physical review E  70(6),  066111 (2004)

\bibitem{pmid22846459}
Das, J., Yu, H.: {{H}{I}{N}{T}: {H}igh-quality protein interactomes and their
  applications in understanding human disease}. BMC Syst Biol  6, ~92 (Jul
  2012)

\bibitem{dimopoulos1995use}
Dimopoulos, Y., Bourret, P., Lek, S.: Use of some sensitivity criteria for
  choosing networks with good generalization ability. Neural Processing Letters
   2(6),  1--4 (1995)

\bibitem{dutkowski2011protein}
Dutkowski, J., Ideker, T.: Protein networks as logic functions in development
  and cancer. PLoS computational biology  7(9),  e1002180 (2011)

\bibitem{pmid17098774}
Falcon, S., Gentleman, R.: {{U}sing {G}{O}stats to test gene lists for {G}{O}
  term association}. Bioinformatics  23(2),  257--258 (Jan 2007)

\bibitem{pmid11773440}
Filardo, E.J., Quinn, J.A., Frackelton, A.R., Bland, K.I.: {{E}strogen action
  via the {G} protein-coupled receptor, {G}{P}{R}30: stimulation of adenylyl
  cyclase and c{A}{M}{P}-mediated attenuation of the epidermal growth factor
  receptor-to-{M}{A}{P}{K} signaling axis}. Mol. Endocrinol.  16(1),  70--84
  (Jan 2002)

\bibitem{goodfellow2016deep}
Goodfellow, I., Bengio, Y., Courville, A.: Deep learning. MIT press (2016)

\bibitem{henaff2015deep}
Henaff, M., Bruna, J., LeCun, Y.: Deep convolutional networks on
  graph-structured data. arXiv preprint arXiv:1506.05163  (2015)

\bibitem{hochreiter2001gradient}
Hochreiter, S., Bengio, Y., Frasconi, P., Schmidhuber, J., et~al.: Gradient
  flow in recurrent nets: the difficulty of learning long-term dependencies
  (2001)

\bibitem{pmid10200323}
Improta-Brears, T., Whorton, A.R., Codazzi, F., York, J.D., Meyer, T.,
  McDonnell, D.P.: {{E}strogen-induced activation of mitogen-activated protein
  kinase requires mobilization of intracellular calcium}. Proc. Natl. Acad.
  Sci. U.S.A.  96(8),  4686--4691 (Apr 1999)

\bibitem{pmid25499220}
Jung, Y.S., Chun, H.Y., Yoon, M.H., Park, B.J.: {{E}levated estrogen
  receptor-Î± in {V}{H}{L}-deficient condition induces microtubule organizing
  center amplification via disruption of {B}{R}{C}{A}1/{R}ad51 interaction}.
  Neoplasia  16(12),  1070--1081 (Dec 2014)

\bibitem{pmid7491495}
Kato, S., Endoh, H., Masuhiro, Y., Kitamoto, T., Uchiyama, S., Sasaki, H.,
  Masushige, S., Gotoh, Y., Nishida, E., Kawashima, H., Metzger, D., Chambon,
  P.: {{A}ctivation of the estrogen receptor through phosphorylation by
  mitogen-activated protein kinase}. Science  270(5241),  1491--1494 (Dec 1995)

\bibitem{kim2013network}
Kim, S., Pan, W., Shen, X.: Network-based penalized regression with application
  to genomic data. Biometrics  69(3),  582--593 (2013)

\bibitem{DBLP:journals/corr/KingmaB14}
Kingma, D.P., Ba, J.: Adam: {A} method for stochastic optimization. CoRR
  abs/1412.6980 (2014), \url{http://arxiv.org/abs/1412.6980}

\bibitem{koboldt2012comprehensive}
Koboldt, D.C., Fulton, R.S., McLellan, M.D., Schmidt, H., Kalicki-Veizer, J.,
  McMichael, J.F., Fulton, L.L., Dooling, D.J., Ding, L., Mardis, E.R., et~al.:
  Comprehensive molecular portraits of human breast tumours. Nature  490,
  61--70 (2012)

\bibitem{Kolaczyk:2009:SAN:1593430}
Kolaczyk, E.D.: Statistical Analysis of Network Data: Methods and Models.
  Springer Publishing Company, Incorporated, 1st edn. (2009)

\bibitem{kursa2014robustness}
Kursa, M.B.: Robustness of random forest-based gene selection methods. BMC
  bioinformatics  15(1), ~8 (2014)

\bibitem{pmid18991569}
Lanzino, M., Morelli, C., Garofalo, C., Panno, M.L., Mauro, L., Ando, S.,
  Sisci, D.: {{I}nteraction between estrogen receptor alpha and
  insulin/{I}{G}{F} signaling in breast cancer}. Curr Cancer Drug Targets
  8(7),  597--610 (Nov 2008)

\bibitem{lavi2012network}
Lavi, O., Dror, G., Shamir, R.: Network-induced classification kernels for gene
  expression profile analysis. Journal of Computational Biology  19(6),
  694--709 (2012)

\bibitem{lecun2015deep}
LeCun, Y., Bengio, Y., Hinton, G.: Deep learning. Nature  521(7553),  436--444
  (2015)

\bibitem{lecun-mnisthandwrittendigit-2010}
LeCun, Y., Cortes, C.: {MNIST} handwritten digit database  (2010),
  \url{http://yann.lecun.com/exdb/mnist/}

\bibitem{pmid15601761}
Li, J.J., Weroha, S.J., Lingle, W.L., Papa, D., Salisbury, J.L., Li, S.A.:
  {{E}strogen mediates {A}urora-{A} overexpression, centrosome amplification,
  chromosomal instability, and breast cancer in female {A}{C}{I} rats}. Proc.
  Natl. Acad. Sci. U.S.A.  101(52),  18123--18128 (Dec 2004)

\bibitem{liang2013sparse}
Liang, Y., Liu, C., Luan, X.Z., Leung, K.S., Chan, T.M., Xu, Z.B., Zhang, H.:
  Sparse logistic regression with a l 1/2 penalty for gene selection in cancer
  classification. BMC bioinformatics  14(1),  198 (2013)

\bibitem{min2016deep}
Min, S., Lee, B., Yoon, S.: Deep learning in bioinformatics. Briefings in
  bioinformatics p. bbw068 (2016)

\bibitem{mockus2012bayesian}
Mockus, J.: Bayesian approach to global optimization: theory and applications,
  vol.~37. Springer Science \& Business Media (2012)

\bibitem{nair2010rectified}
Nair, V., Hinton, G.E.: Rectified linear units improve restricted boltzmann
  machines. In: Proceedings of the 27th international conference on machine
  learning (ICML-10). pp. 807--814 (2010)

\bibitem{ILSVRC15}
Russakovsky, O., Deng, J., Su, H., Krause, J., Satheesh, S., Ma, S., Huang, Z.,
  Karpathy, A., Khosla, A., Bernstein, M., Berg, A.C., Fei-Fei, L.: {ImageNet
  Large Scale Visual Recognition Challenge}. International Journal of Computer
  Vision (IJCV)  115(3),  211--252 (2015)

\bibitem{pmid12829800}
Sorlie, T., Tibshirani, R., Parker, J., Hastie, T., Marron, J.S., Nobel, A.,
  Deng, S., Johnsen, H., Pesich, R., Geisler, S., Demeter, J., Perou, C.M.,
  L?nning, P.E., Brown, P.O., B?rresen-Dale, A.L., Botstein, D.: {{R}epeated
  observation of breast tumor subtypes in independent gene expression data
  sets}. Proc. Natl. Acad. Sci. U.S.A.  100(14),  8418--8423 (Jul 2003)

\bibitem{pmid25859942}
Szklarczyk, D., Jensen, L.J.: {{P}rotein-protein interaction databases}.
  Methods Mol. Biol.  1278,  39--56 (2015)

\bibitem{vanitha2015gene}
Vanitha, C.D.A., Devaraj, D., Venkatesulu, M.: Gene expression data
  classification using support vector machine and mutual information-based gene
  selection. Procedia Computer Science  47,  13--21 (2015)

\bibitem{wei2007incorporating}
Wei, P., Pan, W.: Incorporating gene networks into statistical tests for
  genomic data via a spatially correlated mixture model. Bioinformatics  24(3),
   404--411 (2007)

\bibitem{pmid25275589}
Wei, W., Chen, Z.J., Zhang, K.S., Yang, X.L., Wu, Y.M., Chen, X.H., Huang,
  H.B., Liu, H.L., Cai, S.H., Du, J., Wang, H.S.: {{T}he activation of {G}
  protein-coupled receptor 30 ({G}{P}{R}30) inhibits proliferation of estrogen
  receptor-negative breast cancer cells in vitro and in vivo}. Cell Death Dis
  5,  e1428 (Oct 2014)

\bibitem{pmid18183590}
Zheng, X., Resnick, R.J., Shalloway, D.: {{A}poptosis of estrogen-receptor
  negative breast cancer and colon cancer cell lines by {P}{T}{P} alpha and src
  {R}{N}{A}i}. Int. J. Cancer  122(9),  1999--2007 (May 2008)

\bibitem{zhu2009network}
Zhu, Y., Shen, X., Pan, W.: Network-based support vector machine for
  classification of microarray samples. BMC bioinformatics  10(1),  S21 (2009)

\end{thebibliography}

\end{document}